\documentclass[10pt,twocolumn,letterpaper]{article}

\usepackage{cvpr}

\usepackage{microtype}

\definecolor{cvprblue}{rgb}{0.21,0.49,0.74}
\usepackage[pagebackref,breaklinks,colorlinks,allcolors=cvprblue]{hyperref}

\def\paperID{*****}
\def\confName{CVPR}
\def\confYear{2026}

\newcommand{\myparagraph}[1]{\vspace{.3em}\noindent\textbf{#1.}\;}

\title{WristCompass: Kinematic Coupling as a Learnable Visual Concept
for Ego-Camera Orientation}

\author{%
Varun Nair \quad Vidyut Baradwaj \quad Jiahang He \quad
Anya Singh \quad Jai Relan \quad Cabrel Happi
}

\begin{document}
\maketitle

\begin{abstract}
Recovering ego-camera orientation from manipulation video is a prerequisite for disentangling hand motion from camera motion, a key step in imitation learning from egocentric demonstrations. The obvious approach, inferring orientation from scene geometry, fails when hands occlude the frame: VGGT, a 1B-parameter scene reconstruction model, scores worse than a constant predictor on the TACO benchmark. We identify an alternative visual concept that is present precisely when scene geometry is absent: kinematic coupling dynamics, the structured physical relationship between wrist motion and camera orientation imposed by the arm-shoulder-head chain. We find that this concept is compact (4D inter-wrist features outperform 126D full hand keypoints), temporal (requiring a GRU over short windows rather than per-frame retrieval), and physically grounded (transferring zero-shot across datasets because it is rooted in anatomy rather than scene appearance). Trained only on tabletop manipulation, WristCompass transfers zero-shot to Epic Kitchens cooking video, achieving 14.3° median geodesic error and approaching the performance of a 1B-parameter scene model at 200K GRU parameters.
\end{abstract}

\section{Introduction}
\label{sec:intro}

Egocentric manipulation video is an observation from a joint space:
ego motion (camera orientation in $SO(3)$) entangled with world
state (hand and object configuration). Disentangling these
components requires identifying the right visual concept for each, but the camera orientation component has proven surprisingly hard to
recover. Scene-based approaches fail where
hands occlude the frame: VGGT~\cite{vggt}, a 1B-parameter scene
reconstruction model, scores 23.98$^\circ$ median geodesic error
on the bimanual manipulation benchmark TACO~\cite{taco}, worse
than a constant predictor (21.22$^\circ$). The scene-geometry
concept simply does not exist when the scene is occluded.

We identify the concept that does exist and is sufficient for
recovering the camera orientation: \emph{kinematic coupling
dynamics}. The arm-shoulder-head chain imposes a structured
physical relationship between wrist motion and camera orientation.
This suggests that egocentric video contains a compact, structured
representation of ego motion encoded in body dynamics, a visual
concept that is present precisely when scene geometry is absent.

This concept has three properties that characterize it as a
structured visual concept.
\textbf{Compact:} 4D inter-wrist features outperform 126D full hand keypoints (17.5$^\circ$ vs.\
13.80$^\circ$), consistent with the signal concentrating in the
inter-wrist vector.
\textbf{Temporal:} a nearest-neighbor lookup on per-frame features
scores 26.69$^\circ$, worse than constant, while a GRU over
12-frame windows achieves 13.80$^\circ$; the concept is not
accessible to static retrieval, only to temporal models.
\textbf{Physically grounded:} trained only on TACO tabletop
manipulation, WristCompass transfers zero-shot to Epic
Kitchens~\cite{epickitchens}. It achieves 14.32°, approaching the performance of VGGT at 1000× lower parameter count, because the concept is grounded in anatomy rather than scene appearance.

\noindent\textbf{Contributions: (1)} identifying kinematic coupling
dynamics as a compact, structured, temporally-grounded visual
concept sufficient for camera orientation recovery in manipulation video;
\textbf{(2)} WristCompass, realizing this concept from bare RGB;
\textbf{(3)} a zero-shot transfer result demonstrating that the concept generalizes across datasets because it is grounded in anatomy rather than scene appearance.

\section{Related Work}
\label{sec:related}

\myparagraph{Ego-camera pose estimation}
Classical approaches rely on scene structure: SLAM
systems~\cite{orbslam3} track sparse keypoints across frames, while
COLMAP~\cite{colmap} reconstructs camera trajectories via
structure-from-motion. These methods require sufficient scene
texture and viewpoint diversity, assumptions that break on
close-up manipulation video where hands occlude most of the frame.
VGGT~\cite{vggt}, a 1B-parameter transformer trained for 3D scene
reconstruction, inherits the same failure mode: despite
state-of-the-art performance on standard benchmarks, it scores
23.98$^\circ$ geodesic error on TACO --- worse than a constant
rotation (21.22$^\circ$). IMU-based approaches can recover
orientation but require dedicated hardware unavailable in existing
RGB-only video archives. WristCompass targets precisely this regime
--- recovering orientation post-hoc from bare monocular RGB.

\myparagraph{Ego-body and head pose estimation}
EgoAllo~\cite{egoallo} and EgoEgo~\cite{egoego} recover head or
body pose from egocentric video, but assume a wide-field view in
which the body is partially visible --- an assumption that fails in
close-up manipulation. WristCompass inverts the direction: we
recover camera orientation from wrist dynamics in the ego frame.

\myparagraph{Hand pose and motion estimation}
WiLoR~\cite{wilor} recovers 3D hand meshes from monocular RGB ---
we use it as our keypoint extractor. HaWoR~\cite{Zhang2025HaWoRWH} extends
this to world-space trajectories via SLAM-based tracking. Both
treat hand pose as output requiring camera pose as input.
WristCompass reverses this: wrist geometry predicts camera
orientation. We discuss kinematic coupling in relation to concept
learning methods ($\beta$-VAE~\cite{betavae}, Slot
Attention~\cite{slotattention}, CBMs~\cite{cbm}) in Sec.~5.

\vspace{-0.5em}
\section{Method}
\label{sec:method}

\myparagraph{Input representation}
Given an egocentric video, we extract 3D hand keypoints using
WiLoR~\cite{wilor}. From each frame, we take the two wrist
positions --- right wrist (joint 0) and left wrist (joint 21) ---
and compute a 4D inter-wrist feature vector:
\begin{equation}
\mathbf{f}_t = \left[ \|\mathbf{d}_t\|,\;
\frac{\mathbf{d}_t}{\|\mathbf{d}_t\|} \right] \in \mathbb{R}^4
\label{eq:features}
\end{equation}
where $\mathbf{d}_t = \mathbf{w}^L_t - \mathbf{w}^R_t$ is the
left-minus-right wrist difference vector. The first component is
inter-wrist distance; the remaining three form a unit direction
vector on $S^2$. Features are z-score normalized using training
set statistics and mean-centred per video at test time (using
per-video statistics computed from the test video itself) to remove
postural bias without leaking cross-video information.

\myparagraph{Why 4D beats 126D}
A full hand keypoint representation (42 joints $\times$ 3D = 126D)
achieves 17.5$^\circ$ on TACO with an MLP --- worse than our 4D
representation (13.80$^\circ$). To control for architecture, we
also train an identical GRU (same hidden size, window, optimizer)
on 126D input: 4D outperforms 126D on all 5 folds of a
subject-level cross-validation. The orientation signal
concentrates in the inter-wrist vector; additional finger
articulations add subject-specific noise that hurts generalization
to held-out subjects.

\myparagraph{Temporal model}
We process the feature sequence with a two-layer GRU:
\begin{equation}
\mathbf{h}_t = \mathrm{GRU}(\mathbf{f}_{t-W:t};\,\theta),
\quad W = 12
\label{eq:gru}
\end{equation}
where $W{=}12$ frames (${\approx}0.4$s at 30fps). A linear head
maps $\mathbf{h}_t \in \mathbb{R}^{128}$ to a 6D rotation
representation~\cite{zhou2019}, projected to $SO(3)$ via
Gram-Schmidt orthogonalization. We train by minimizing geodesic
loss against TACO ground-truth rotations from NOKOV optical motion
capture, using Adam ($\mathrm{lr}{=}5{\times}10^{-4}$) with
early stopping on a held-out validation split (frame-level 80/20,
used solely for early stopping).

\myparagraph{Inference}
WristCompass has 200K parameters (excluding WiLoR as a shared
frozen feature extractor). At inference, we run WiLoR on each
frame and compute $\mathbf{f}_t$ when both wrists are detected;
frames with only one detected wrist are dropped (not interpolated).
On Epic Kitchens (${\approx}$50\% bimanual detection), evaluation
is computed over bimanual frames only; GRU windows span
non-consecutive frames when detections are sparse.
We pass a sliding window of 12 frames to the GRU. The full
pipeline runs in real time on CPU from bare monocular RGB.
Evaluation uses Procrustes alignment --- a single global rotation
applied to all predictions within a video to minimize mean geodesic
error --- measuring relative orientation structure rather than
absolute pose.

\begin{figure*}[t]
  \centering
  \includegraphics[width=\linewidth]{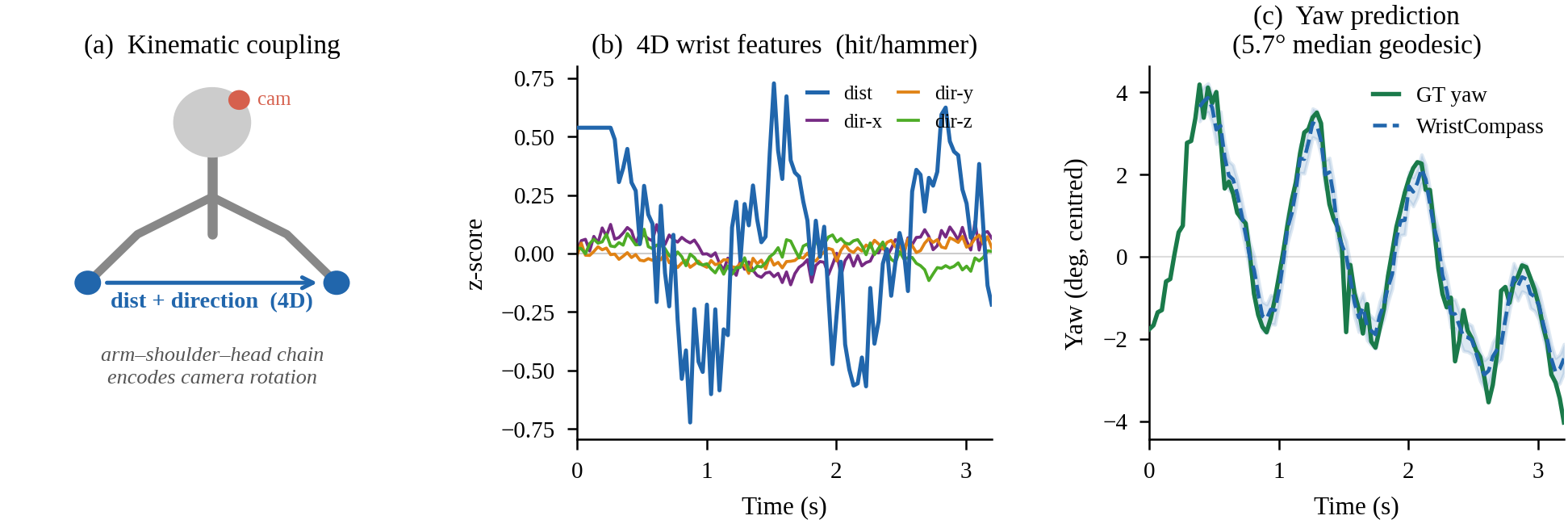}
  \caption{\textbf{WristCompass overview.}
  (a)~Kinematic coupling: the arm-shoulder-head chain couples wrist
  motion dynamics to ego-camera rotation.
  (b)~4D inter-wrist features for a representative session --- only
  the distance feature carries temporal variation at this timescale;
  direction components encode relative hand geometry.
  (c)~WristCompass predicted yaw vs.\ ground-truth (GT) yaw
  (5.7$^\circ$ median geodesic). Shaded region indicates prediction
  uncertainty. The model tracks head rotation from wrist dynamics
  alone, with no scene information.}
  \label{fig:teaser}
\end{figure*}

\vspace{-0.5em}
\section{Experiments}
\label{sec:experiments}

\myparagraph{Datasets and evaluation}
\textbf{TACO}~\cite{taco} provides 17 subjects performing 15
tool-action-object activities (5,210 frames) with a helmet-mounted
RealSense L515 and NOKOV optical motion-capture ground truth for
camera rotation. We report median geodesic error after Procrustes
alignment, averaged over 5 random seeds on a fixed train/val split
(80/20, stratified by subject), with early stopping on the
held-out portion.

\textbf{Epic Kitchens}~\cite{epickitchens} is a large-scale
egocentric cooking dataset with a chest-mounted GoPro. We use
EPIC Fields~\cite{epicfields} COLMAP poses as ground truth and
evaluate on 36 participants, 62 videos, 16,609 frames (minimum
thresholds: 70\% COLMAP coverage, 10$^\circ$ constant baseline per
video). COLMAP poses are a proxy for ground truth and may exhibit
drift in textureless or heavily occluded regions.

\subsection{TACO In-Distribution Results}

\begin{figure}[t]
  \centering
  \includegraphics[width=\linewidth]{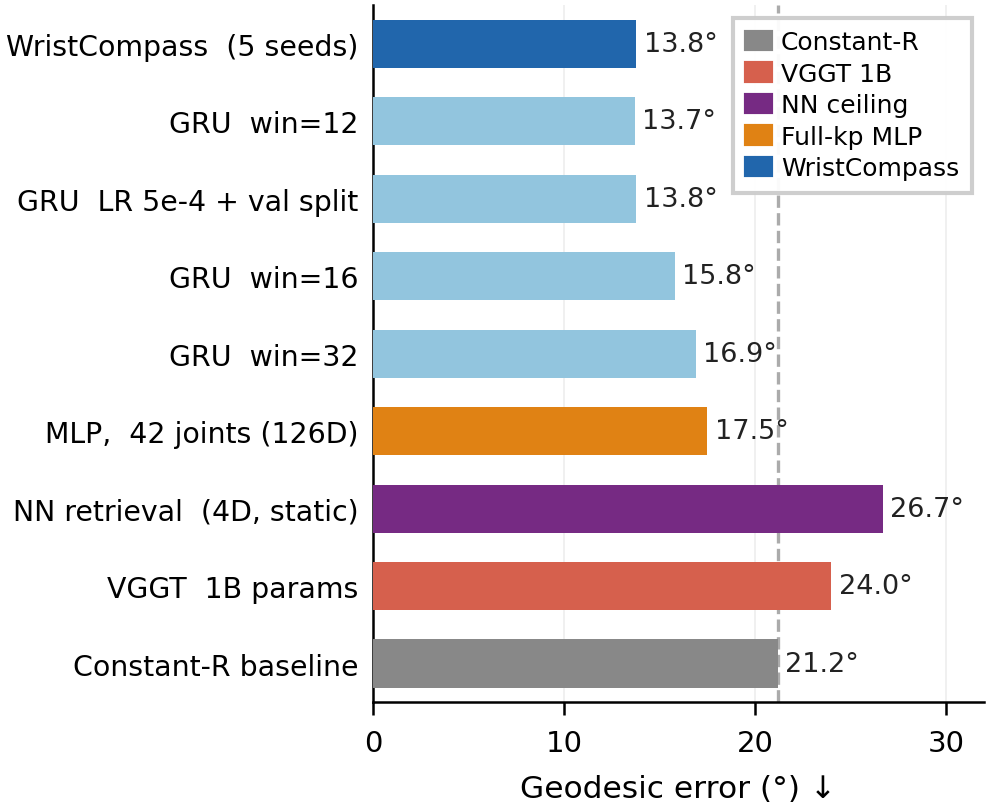}
  \caption{\textbf{TACO ablation.} NN retrieval (26.7$^\circ$) is
  worse than the constant baseline (21.2$^\circ$) --- the
  orientation signal is not accessible to static retrieval.
  WristCompass (13.8$^\circ$) outperforms VGGT 1B (24.0$^\circ$)
  at 200K GRU parameters.}
  \label{fig:ablation}
\end{figure}

Table~\ref{tab:ablation} and Figure~\ref{fig:ablation} show the
full ablation. WristCompass achieves
\textbf{13.80$^\circ$\,$\pm$\,0.14$^\circ$}, outperforming all
baselines.

Three findings stand out. \textbf{First}, VGGT (23.98$^\circ$) is
worse than the constant predictor (21.22$^\circ$) --- confirming
that scene-based approaches fail on close-up manipulation video.
\textbf{Second}, the 126D full-keypoint MLP (17.5$^\circ$) is
worse than 4D wrist geometry. A controlled comparison --- identical
GRU architecture on 126D input --- confirms: 4D wins on all 5
folds of subject-level cross-validation, ruling out architecture
and capacity as confounds.
\textbf{Third}, NN retrieval on the same 4D features scores
26.69$^\circ$, worse than constant --- the orientation signal is
not accessible to static retrieval, requiring temporal context.

\begin{table}[t]
\centering
\small
\setlength{\tabcolsep}{6pt}
\caption{\textbf{TACO ablation.} All GRU rows use 4D inter-wrist
features unless noted. A controlled comparison (same GRU on 126D
input) confirms 4D wins on all 5 folds of subject-level CV (see
Sec.~3). Lower is better.}
\label{tab:ablation}
\begin{tabular}{lc}
\toprule
Method & Geodesic ($^\circ$) $\downarrow$ \\
\midrule
Constant-R baseline            & 21.2 \\
VGGT 1B~\cite{vggt}           & 24.0 \\
NN retrieval (4D, static)      & 26.7 \\
MLP, 42 joints (126D)          & 17.5 \\
\midrule
GRU $W{=}32$ (init)            & 16.9 \\
\quad $\to$ $W{=}16$           & 15.8 \\
\quad $\to$ LR $5{\times}10^{-4}$ + val split & 13.8 \\
\quad $\to$ $W{=}12$           & 13.7 \\
\midrule
\textbf{WristCompass} (5 seeds) & \textbf{13.8\,$\pm$\,0.14} \\
\bottomrule
\end{tabular}
\end{table}

\begin{figure}[t]
  \centering
  \includegraphics[width=\linewidth]{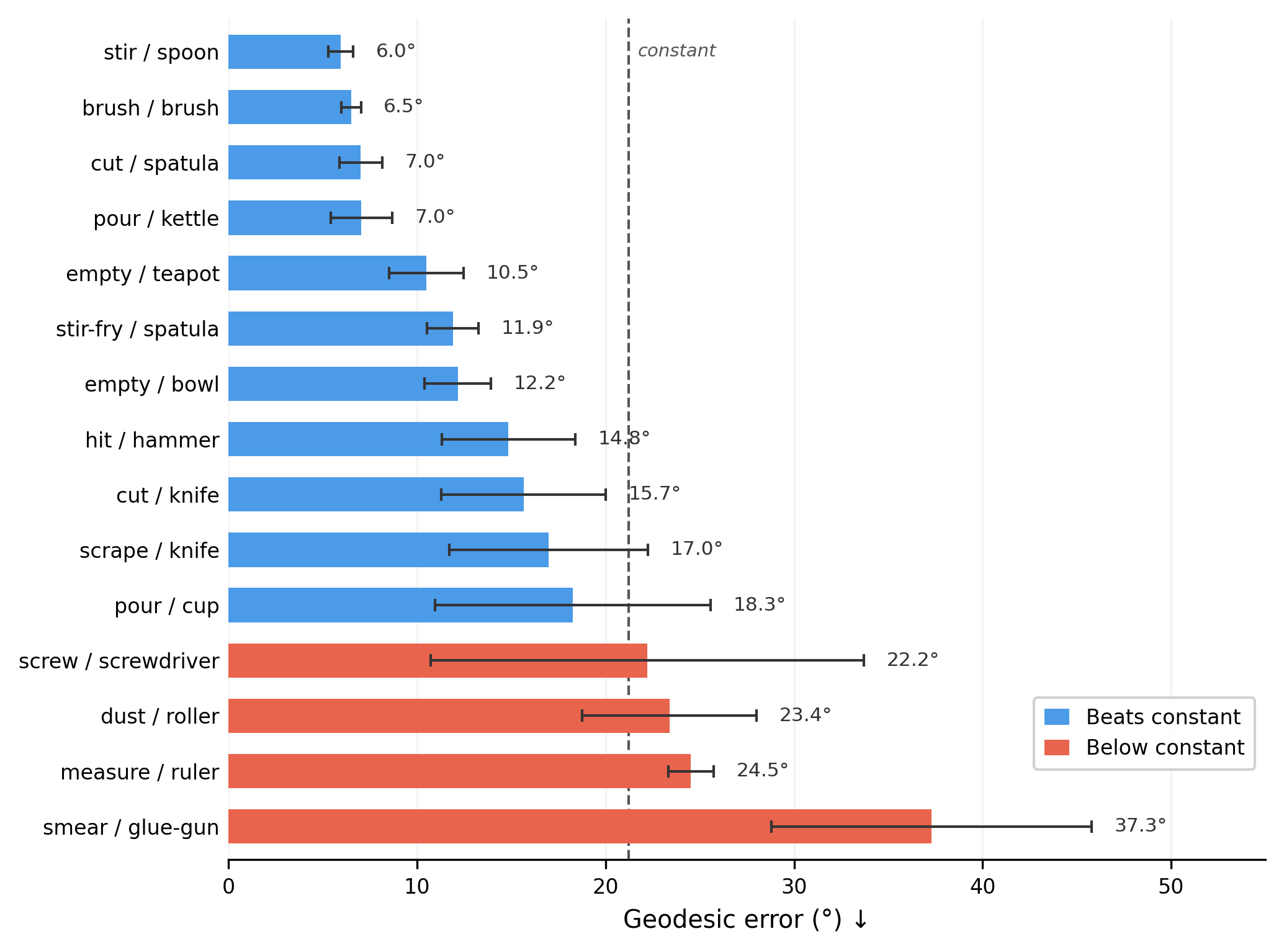}
  \caption{\textbf{Per-activity TACO results} (5 seeds, IQR
  error bars). Blue: beats constant (11/15). Red: below constant
  (4/15). Cyclic motions (stir, brush) are easiest; decoupled
  motions (smear, measure) are hardest.}
  \label{fig:activity}
\end{figure}

\myparagraph{Per-activity analysis}
Figure~\ref{fig:activity} shows per-activity results. WristCompass
beats the constant baseline on 11/15 activities. Best on cyclic
motions: stir/spoon (5.9$^\circ$), brush/brush (6.5$^\circ$).
Worst on activities where head orientation decouples from wrist
dynamics: smear/glue-gun (37.3$^\circ$), measure/ruler
(24.3$^\circ$) --- in these cases, the subject's wrists remain
nearly stationary while the head rotates to inspect different
workspace regions, breaking the kinematic coupling assumption.

\subsection{Zero-Shot Transfer to Epic Kitchens}

\begin{figure}[t]
  \centering
  \includegraphics[width=\linewidth]{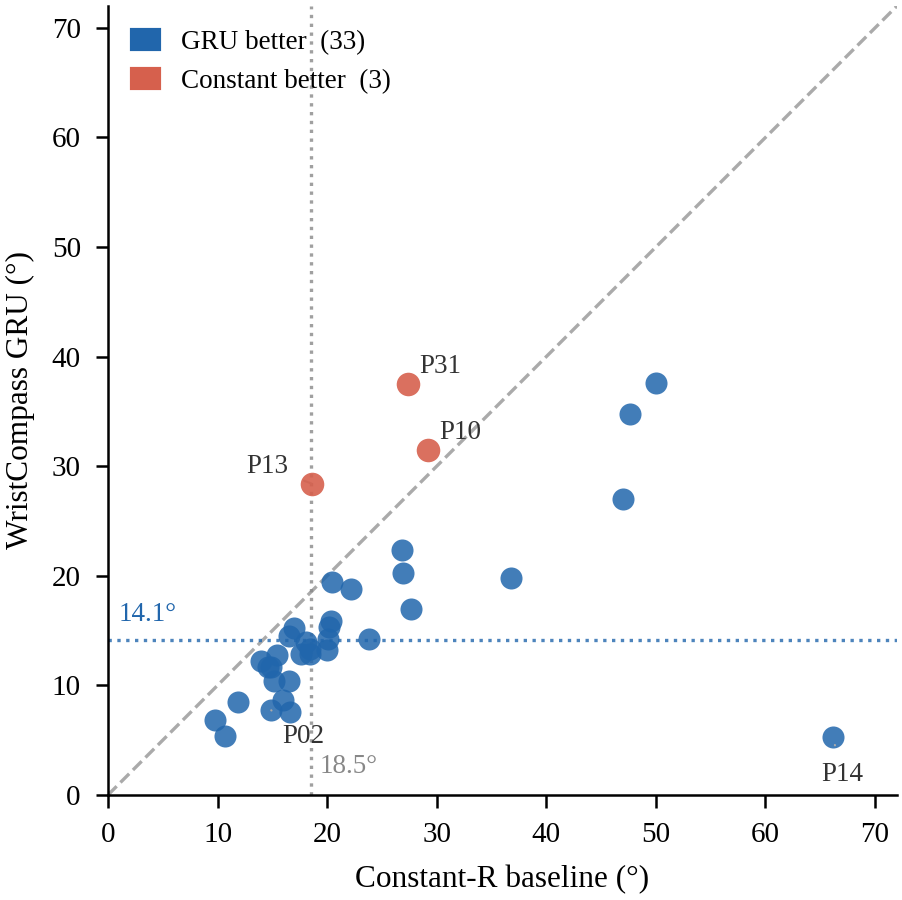}
  \caption{\textbf{Epic Kitchens zero-shot (36 participants).}
  Points below diagonal: WristCompass beats constant baseline
  (33/36, blue). Points above: failures (3/36, red). P14 (constant
  66$^\circ$, GRU 5$^\circ$) is a striking outlier.}
  \label{fig:scatter}
\end{figure}

Trained exclusively on TACO, WristCompass achieves
\textbf{14.32$^\circ$} zero-shot on Epic Kitchens --- 33/36
participants beat the constant baseline.
Despite different ground-truth sources (mocap vs.\ COLMAP), camera
mountings (helmet vs.\ chest), and activity domains, performance is
comparable to in-distribution TACO (13.80$^\circ$).
Figure~\ref{fig:scatter} shows the per-participant breakdown.
P14 (constant 66.1$^\circ$ $\to$ GRU 5.3$^\circ$) demonstrates
strong signal extraction when head movement is rich. The three
failures (P10, P13, P31) are single-video participants with limited
COLMAP pose quality.

\subsection{WristCompass vs.\ VGGT on Epic Kitchens}

\begin{table}[t]
\centering
\small
\setlength{\tabcolsep}{6pt}
\caption{\textbf{Epic Kitchens comparison} (36 participants,
zero-shot for WristCompass). Parameter count for WristCompass
refers to the GRU only; WiLoR keypoint extractor is a shared
prerequisite not counted in this comparison.}
\label{tab:vggt}
\begin{tabular}{lcc}
\toprule
Method & Params & Geodesic ($^\circ$) $\downarrow$ \\
\midrule
Constant-R           & ---   & 18.5 \\
VGGT~\cite{vggt}    & 1B    & 12.8 \\
\textbf{WristCompass} & \textbf{200K} & \textbf{14.3} \\
\bottomrule
\end{tabular}
\end{table}

VGGT achieves 12.83$^\circ$ vs.\ WristCompass 14.32$^\circ$ ---
a 1.5$^\circ$ gap at 1000$\times$ lower GRU parameter count (COLMAP
ground truth may favor VGGT, which shares its scene-geometry
assumptions). Both substantially beat the constant baseline
(18.54$^\circ$). The methods have complementary failure modes:
VGGT wins on discrete manipulation tasks (cut, pick-up, take)
where scene structure changes predictably, while WristCompass wins
on cyclic tasks (stir) where inter-wrist dynamics are more stable
than dynamic scene features. On TACO --- where scene features are
largely occluded by close-up hand manipulation --- WristCompass wins
by over 10$^\circ$.

\vspace{-0.5em}
\section{Discussion and Limitations}
\label{sec:discussion}

\myparagraph{When it works and when it fails}
WristCompass works best when head movement is rich (constant
baseline ${>}$\,15$^\circ$) and both hands are consistently
visible --- P14 (constant 66$^\circ$, GRU 5$^\circ$) shows the
potential ceiling. Cyclic tasks (stir, brush) are particularly
well-suited. The model cannot improve on the constant baseline
when head orientation is near-static: HOT3D (5.1$^\circ$) and
ARCTIC (5.5$^\circ$) both fail this minimum variance threshold.
Activity-level failures (smear 37.3$^\circ$, measure 24.3$^\circ$)
occur when wrist position is constrained independently of head
orientation, breaking kinematic coupling. Both evaluation domains
involve standing manipulation; transfer to seated or full-body
activities remains untested. Procrustes alignment measures relative
orientation structure; absolute calibration is a separate problem.
Relative orientation is nonetheless useful for downstream tasks
such as ego-motion-compensated hand trajectories in imitation
learning, where action representations depend on frame-to-frame
rotation rather than global pose.
Epic Kitchens evaluation uses raw WiLoR-mini keypoints without
smoothing (${\approx}$50\% bimanual detection); Kalman--RTS
smoothing improves TACO (${\approx}$100\% detection) but degrades
Epic Kitchens by replacing temporal signal with near-constant
interpolations.

\myparagraph{Kinematic coupling as a physical visual concept}
Data-driven concept discovery --- $\beta$-VAE~\cite{betavae}, Slot
Attention~\cite{slotattention}, Concept Bottleneck
Models~\cite{cbm} --- learns representations from co-occurrence
statistics, with generalization bounded by training distribution
diversity. Inductive-bias approaches impose architectural priors
(competition, capacity limits) rather than physical
ones~\cite{locatello2019}. Kinematic coupling occupies a third
position: a \emph{physically grounded} concept bottleneck whose
4D inter-wrist representation is dictated by biomechanics, not
learned from data. It is also \emph{intrinsically temporal} ---
the NN ablation shows the concept is not accessible via static
retrieval on these features. The
zero-shot transfer from TACO to Epic Kitchens provides empirical
evidence: the concept transfers because anatomy is shared, not
because the training distribution covers the test distribution.

\vspace{-0.5em}
\section{Conclusion}
\label{sec:conclusion}

We identify kinematic coupling dynamics --- the temporal relationship
between bimanual wrist motion and ego-camera orientation imposed by
the arm-shoulder-head chain --- as a compact, physically-grounded
visual concept for recovering the ego $SO(3)$ component of
manipulation video. WristCompass realizes this concept with 200K
GRU parameters from bare monocular RGB, outperforming a
1B-parameter scene model on close-up manipulation and generalizing
zero-shot to kitchen video. Future directions include depth
integration, explicit ego-world disentanglement~\cite{egoexo4d},
and downstream robot policy learning.

{
    \small
    \bibliographystyle{ieeenat_fullname}
    \bibliography{wristcompass}
}

\clearpage
\setcounter{section}{0}
\renewcommand{\thesection}{\Alph{section}}
\section*{Supplementary Material}

\myparagraph{A.\; Controlled 4D vs.\ 126D comparison}
Table~\ref{tab:controlled} reports a head-to-head comparison
using identical GRU architectures ($W{=}12$, hidden=128,
2 layers) on 4D inter-wrist vs.\ 126D full-keypoint input,
evaluated via 5-fold subject-level cross-validation. 4D
outperforms 126D on all 5 folds, ruling out architecture
and capacity as confounds. The 126D model overfits to
subject-specific finger articulations that do not transfer
to held-out subjects.

\begin{table}[h]
\centering
\small
\setlength{\tabcolsep}{8pt}
\caption{\textbf{4D vs.\ 126D GRU} (5-fold subject-level CV,
5 seeds per fold). 4D wins on every fold.}
\label{tab:controlled}
\begin{tabular}{lccc}
\toprule
Fold & 4D ($^\circ$) & 126D ($^\circ$) & $\Delta$ \\
\midrule
0 & 4.67 & 6.81  & $-$2.14 \\
1 & 6.44 & 7.98  & $-$1.54 \\
2 & 3.68 & 7.82  & $-$4.14 \\
3 & 8.56 & 12.14 & $-$3.58 \\
4 & 2.68 & 6.66  & $-$3.98 \\
\midrule
\textbf{Mean} & \textbf{4.77\,$\pm$\,0.24} & \textbf{7.84\,$\pm$\,0.25} & $-$\textbf{3.07} \\
\bottomrule
\end{tabular}
\end{table}

\myparagraph{B.\; Per-axis error decomposition}
On TACO, WristCompass achieves per-axis median errors of
yaw 8.44$^\circ$, pitch 5.28$^\circ$, roll 3.87$^\circ$.
Yaw (left-right head rotation) carries the most error,
consistent with the inter-wrist vector being most
informative about lateral head movement. Pitch and roll
are better constrained by the arm-shoulder-head kinematic
chain.

\myparagraph{C.\; All-frames Epic Kitchens evaluation}
The main paper reports 14.32$^\circ$ on bimanual frames
only (${\approx}$52\% of frames). For a deployment-realistic
estimate, we blend GRU predictions on bimanual frames with
the constant-R baseline on single-hand/no-hand frames:

\begin{table}[h]
\centering
\small
\setlength{\tabcolsep}{6pt}
\caption{\textbf{Epic Kitchens blended evaluation} (62 videos,
16,609 total frames).}
\label{tab:blended}
\begin{tabular}{lc}
\toprule
Metric & Geodesic ($^\circ$) $\downarrow$ \\
\midrule
Constant-R (all frames)       & 18.54 \\
GRU-only (bimanual, 52\%)    & 14.32 \\
Blended median (per-video)    & 16.56 \\
\bottomrule
\end{tabular}
\end{table}

\noindent The 2.0$^\circ$ gap between GRU-only and blended
reflects the 48\% of frames where only one or no hands are
detected. Improving single-hand fallback is a clear
direction for future work.

\begin{figure}[h]
  \centering
  \includegraphics[width=\linewidth]{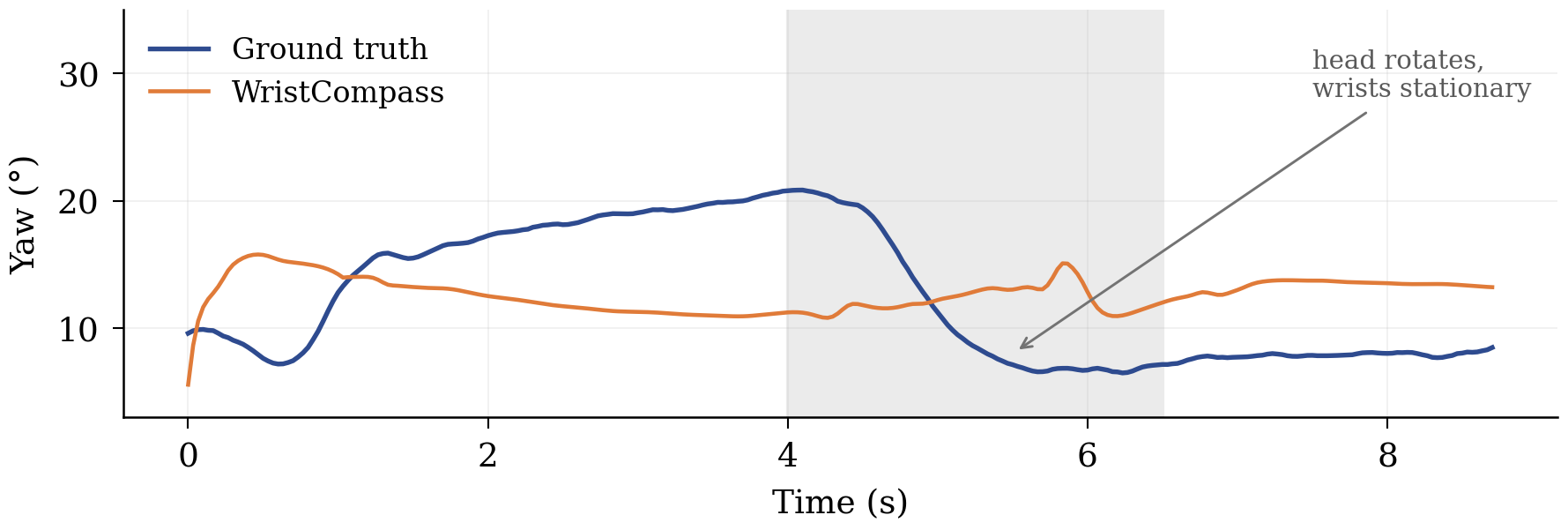}
  \caption{\textbf{Failure case: kinematic decoupling.}
  GT yaw (blue) drops 14$^\circ$ between $t{=}4$--6s while
  WristCompass prediction (orange) remains flat. The subject's
  head rotates independently of their stationary wrists.}
  \label{fig:failure}
\end{figure}

\myparagraph{D.\; Failure case: kinematic decoupling}
Figure~\ref{fig:failure} shows a representative failure
(empty/bowl/plate, session 8, 21.3$^\circ$ geodesic).
Between $t{=}4$--6s, the subject's head rotates ${\approx}$14$^\circ$
in yaw while wrists remain stationary --- the subject looks
away from their hands to inspect the workspace. WristCompass
cannot track this head rotation because the kinematic coupling
between wrist motion and head orientation is broken. This
failure mode is systematic: it accounts for the worst
per-activity results (smear/glue-gun 37.3$^\circ$,
measure/ruler 24.3$^\circ$) where wrist position is
constrained independently of gaze direction.

\end{document}